\documentclass[preprint,12pt]{elsarticle}

\usepackage{amssymb}

\usepackage{amsmath,amsfonts}
\usepackage[ruled,linesnumbered, boxed]{algorithm2e}
\usepackage{graphicx}
\usepackage{multirow}
\usepackage{booktabs}
\usepackage{balance}
\usepackage{subfigure}
\usepackage{color}
\usepackage{mathtools}
\usepackage{amssymb}
\usepackage{tabulary}
\usepackage{booktabs}
\usepackage{mathtools}
\usepackage{amssymb}
\usepackage{tabulary}
\usepackage{booktabs}
\usepackage{url}
\usepackage{natbib}

\newcommand{\etal}{\emph{et al.}\xspace}
\newcommand{\eg}{\emph{e.g.,}\xspace}
\newcommand{\ie}{\emph{i.e.,}\xspace}

\usepackage{color}

\newtheorem{myDef}{Definition}

\journal{Information Sciences}

\begin{document}

\begin{frontmatter}



\title{Mitigating the Performance Sacrifice in DP-Satisfied Federated Settings through Graph Contrastive Learning}


\author[1]{Haoran Yang}
\ead{haoran.yang-2@student.uts.edu.au}
\author[2]{Xiangyu Zhao}
\ead{xianzhao@cityu.edu.hk}
\author[3]{Muyang Li}
\ead{muli0371@uni.sydney.edu.au}
\author[4]{Hongxu Chen}
\ead{hongxu.chen@uq.edu.au}
\author[1]{\\Guandong Xu\corref{cor1}}
\ead{guandong.xu@uts.edu.au}

\address[1]{University of Technology Sydney, NSW, Australia}
\address[2]{City University of Hong Kong, Hong Kong SAR, China}
\address[3]{The University of Sydney, NSW, Australia}
\address[4]{The University of Queensland, QLD, Australia}
\cortext[cor1]{Corresponding author.}

\begin{abstract}
Currently, graph learning models are indispensable tools to help researchers explore graph-structured data. In academia, using sufficient training data to optimize a graph model on a single device is a typical approach for training a capable graph learning model. Due to privacy concerns, however, it is infeasible to do so in real-world scenarios. Federated learning provides a practical means of addressing this limitation by introducing various privacy-preserving mechanisms, such as differential privacy (DP) on the graph edges. However, although DP in federated graph learning can ensure the security of sensitive information represented in graphs, it usually causes the performance of graph learning models to degrade. In this paper, we investigate how DP can be implemented on graph edges and observe a performance decrease in our experiments. In addition, we note that DP on graph edges introduces noise that perturbs graph proximity, which is one of the graph augmentations in graph contrastive learning. Inspired by this, we propose leveraging graph contrastive learning to alleviate the performance drop resulting from DP. Extensive experiments conducted with four representative graph models on five widely used benchmark datasets show that contrastive learning indeed alleviates the models' DP-induced performance drops. 
\end{abstract}




\end{frontmatter}


\section{Introduction}
With the rapid development of the internet and its applications, online activities are producing a massive amount of data in various forms, such as text, images, and graphs, among which graph data, in particular, have recently attracted increasing attention and interest from the research community due to their powerful ability to illustrate the complex relations among different entities \cite{gcn, gat}. To extract the informative semantics contained in graphs, researchers have developed various graph representation learning methods, including graph embedding methods \cite{deepwalk} and graph neural networks (GNNs) \cite{gcn, gat}, which have been utilized in various real-world applications \cite{knowledge-graph, knowledge-survey}. Usually, the basic versions of such methods need to be trained on massive amounts of training data in a centralized manner. However, such a training setting is infeasible in real-world scenarios due to privacy concerns, commercial competition, and increasingly strict regulations and laws \cite{fedml, fedgraphnn}, such as the GDPR\footnote{https://gdpr-info.eu/} and the CCPA\footnote{https://oag.ca.gov/privacy/ccpa}. To address privacy concerns, efforts have been made to introduce federated learning (FL) \cite{fedgraphnn} into the graph learning domain. FL is a distributed learning paradigm that can serve as a suitable solution for many real-world applications, such as the Internet of Things (IoT) \cite{iot}. It enhances privacy by allowing graph models to be trained on local devices while incorporating various privacy-preserving mechanisms, such as differential privacy (DP) on graphs \cite{dp-survey}.

DP is one of the most widely used privacy-preserving methods with a strong guarantee~\cite{dp-def}. The basic idea of DP is to introduce a controlled level of noise into the query results to perturb the consequences of comparing two adjacent datasets~\cite{dp-proof}.
The advantages of DP lie in its ease of practical implementation and its solid theoretical guarantee. It has been proven that the introduction of noise that follows specific Laplacian or Gaussian distributions can achieve privacy protection at a designated level in accordance with the Laplacian or Gaussian mechanism \cite{dp-proof}. This property means that the DP mechanism is robust against any postprocessing and can never be compromised by any algorithms \cite{dp-book}. In the graph learning domain, there are numerous perspectives from which DP can be implemented, such as node privacy, edge privacy, out-link privacy, and partition privacy \cite{dp-challenge}. Among these, edge privacy provides meaningful privacy protection and is extensively utilized in many real-world graph learning applications \cite{dp-survey}. 

Although DP can ensure the security of the sensitive information contained in graph structures and possesses theoretically proven advantages, it always causes the performance of graph learning models to degrade \cite{performance-dropping}. Some previous research efforts have sought to alleviate this performance drop. One representative approach is to generate specific perturbed data and train a model on both clean and perturbed data simultaneously to help the model gain robustness against DP noise \cite{improve}. However, there are two significant limitations to adopting this strategy in federated graph learning. First, this strategy is designed for deep neural networks (DNNs) and image classification tasks. It cannot be directly applied to address graph learning problems with GNNs. The reason for this limitation is that existing DP mechanisms are usually designed for application to tabular data (including images, as an image can be regarded as a table in which each entry is a pixel). The noise produced by a DP mechanism can be easily added to the existing value of each data entry in a table. However, utilizing such a DP mechanism to process graph structures is not straightforward, as there are no explicit tabular data. Second, model training typically requires clean data. However, if DP is utilized to process the data uploaded from edge devices, it is difficult for a model stored on a cloud server to acquire clean data for such a training process. Therefore, a new solution is highly desirable to overcome the above challenges in federated graph learning.

\begin{figure*}[]
	\centering
	\includegraphics[width=0.48\textwidth]{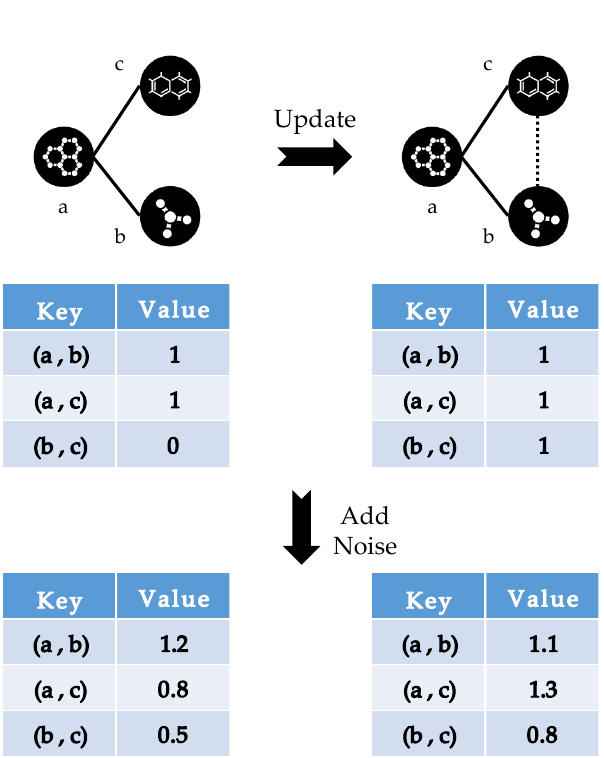}
\caption{After converting the target graph into a table, we utilize the widely used Laplacian mechanism to introduce noise into the values in the table to defend against differential attacks via table lookup.}
	\label{fig:example}
\end{figure*}

To mitigate the performance sacrifice caused by the DP mechanism in federated graph learning \cite{improve}, we aim to develop a solution that is resistant to DP noise. First, we note that a DP mechanism acting on graph edges introduces noise to perturb the graph proximity by interfering with the values of the entries in the adjacency matrix, as shown in Figure \ref{fig:example}. Such an operation can be regarded as a type of graph augmentation, \ie edge perturbation \cite{graphcl}. In addition, we note that You \etal \cite{graphcl} proposed leveraging graph contrastive learning (GCL) \cite{graphcl, dsgc, mvgrl, gcl-attack} in conjunction with graph augmentation to help a graph learning model achieve robustness against noise. Specifically, GCL employs concepts of contrastive learning to extract critical semantics between augmented graphs, where the extracted semantic information is robust to the noise introduced by graph augmentation operations such as edge perturbation \cite{graphcl}. Inspired by the above considerations, we propose utilizing GCL to learn from a graph processed with a DP mechanism.


Nevertheless, applying GCL in federated graph learning is challenging and not straightforward.
First, the noise introduced by DP follows specific distributions (\eg a Laplacian distribution), indicating that simply deleting or adding edges via GCL techniques is insufficient to satisfy a specific form of DP. Moreover, edge perturbation refers to the operation of deleting or adding complete edges in a graph, changes that are reflected by binary entry values of $\{0, 1\}$ in the adjacency matrix. However, the DP mechanism produces noise values in the range of $(0, 1)$.
To ensure compatibility with the graph learning setting, we need to convert the graph of interest into a fully connected graph and use the perturbed probabilities to represent the edge weights.
Additionally, in real-world settings, federated frameworks typically optimize models in a streaming manner, which means that the training batch size is 1.
Consequently, it is not possible to apply the conventional contrastive learning protocol, which requires negative samples to be collected in the same training batch \cite{graphcl, dsgc}, as per the settings in the current GCL literature.
In this paper, we maintain a limited-size stack on each device to store previously trained graphs and randomly extract samples from this stack to serve as negative samples in each training round.

To help better understand the mechanism for achieving privacy preservation on graph structures in our proposed method, we give an illustrative example here, as shown in Fig. \ref{fig:example}. Let us assume that there is a network consisting of three participants. Initially, $a$ has connections with both $b$ and $c$. Then, a new connection is built between $b$ and $c$. If a malicious attacker initiates a differential attack on this social network in two different time slots, he or she will be aware that there has been a change in the relationship between $b$ and $c$, which could be private information. To defend against such attacks, we introduce noise $Y\sim L(0, \frac{\Delta f}{\epsilon})$ to perturb the values, thereby preventing the attacker from being able to determine whether any two participants are connected or not via single or several differential attacks. Because the expectation value of the introduced noise is $0$, many queries and average look-up results over a short period of time would be required to derive the precise lookup result, which is infeasible for the malicious attacker.

In summary, the contributions of this paper are as follows:
\begin{itemize}
\item We propose a method for converting a graph into a table to facilitate the use of current DP methods to achieve privacy preservation on graphs.
\item A novel federated graph contrastive learning method (FGCL) is proposed for the first time. The proposed method can be implemented in a plug-in manner for federated graph learning with any graph encoder.
\item Extensive experiments show the proposed method can alleviate the performance decline caused by the noise introduced by a DP mechanism.
\end{itemize}

\section{Preliminaries}
This section presents some background knowledge related to the current research work, including DP and GCL.

\subsection{Differential Privacy (DP)}
DP refers to a tailored mechanism for privacy-preserving data analysis \cite{improve}, which is designed to defend against differential attacks \cite{dp-book}. DP offers a theoretically proven guarantee that the amount of information leakage that occurs during a malicious attack can be controlled to a certain level \cite{dp-challenge}. This property means that the DP mechanism is immune to any postprocessing and cannot be compromised by any algorithms \cite{dp-book}. In other words, if a dataset is protected by the DP mechanism, the privacy loss suffered by this dataset during a malicious attack will never exceed the specified threshold, even if it is attacked using the most advanced and sophisticated methods available.

Formally, DP is defined as follows:
\begin{myDef}
\cite{dp-book} A randomized query algorithm $\mathcal{A}$ satisfies $\epsilon$-differential privacy iff for any datasets $d$ and $d'$ that differ by a single element and any output $S$ of $\mathcal{A}$,
\begin{equation}
		Pr[\mathcal{A}(d)=S]\leq e^{\epsilon}\cdot Pr[\mathcal{A}(d')=S].
	\end{equation}
\end{myDef}
In the definition above, the datasets $d$ and $d'$ that differ by only one single data item are called adjacent datasets, and this adjacency is defined in a task-specific manner. For example, in graph edge privacy, a graph $G$ represents a dataset, and a graph $G'$ is an adjacent dataset of $G$ if $G'$ can be derived from $G$ by adding or deleting only one edge. The parameter $\epsilon$ in the definition above denotes the privacy budget \cite{dp-book}, which controls how much privacy leakage by $\mathcal{A}$ can be tolerated. A smaller privacy budget indicates less tolerance to information leakage, which is equivalent to a higher level of privacy.

A general method for converting a deterministic query function $f$ into a randomized query function $\mathcal{A}_f$ is to add random noise to the output of $f$. The noise is generated from a specific random distribution calibrated to the privacy budget and $\Delta f$, the global sensitivity of $f$, defined as the maximal value of $||f(d)-f(d')||$. In this paper, for simplicity, we consider the Laplacian mechanism to achieve DP \cite{dp-book}:
\begin{equation}
	\mathcal{A}_f(d) = f(d)+Y,
\end{equation}
where $Y\sim L(0, \frac{\Delta f}{\epsilon})$, with $L(\cdot, \cdot)$ denoting a Laplacian distribution with two parameters to be determined.

\subsection{Graph Augmentation for Graph Contrastive Learning (GCL)}
GCL \cite{graphcl, dsgc, dgi, gca, cgc} has emerged as a fine tool for learning high-quality graph representations. Graph augmentation techniques contribute greatly to the success of GCL, playing a critical role in the GCL process, and many researchers have investigated the implementation of such techniques.
GraphCL \cite{graphcl} is one of the most impactful works in the GCL research community, providing comprehensive and insightful analysis of various graph augmentation operations. In practice, four basic graph augmentation methods and their corresponding underlying priors can be identified, as summarized below:
\begin{itemize}
\item \textbf{Node Drooping.} A small proportion of vertices can be deleted such that the semantics of the modified graphs do not change.
\item \textbf{Edge Perturbation.} Graphs usually exhibit semantic robustness against limited-scale proximity variations.
\item \textbf{Attribute Masking.} Graphs exhibit semantic robustness to a partial loss of attributes on their nodes or edges or on the graphs themselves.
\item \textbf{Subgraph Sampling.} Local subgraphs can hint at the entire semantics of the original graph.
\end{itemize}
In summary, the intuitive purpose behind graph augmentation is to introduce noise into a graph. The augmented graphs can subsequently be utilized in GCL training procedures to help endow a model with the semantic robustness of the original graph. This intuition provides researchers with a potentially feasible way to bridge graph-level privacy preservation and graph augmentation, as well-designed graph augmentation techniques can satisfy the requirements of a DP mechanism.

The graph augmentation methods mentioned above are randomly implemented, achieving suboptimal performance for GCL in practice \cite{gca}. GCA \cite{gca} is an improved version of GraphCL in which adaptive augmentations are applied to achieve better performance. As an alternative to the aforementioned noise-based graph augmentations, other methods, such as MVGRL \cite{mvgrl} and DSGC \cite{dsgc}, alleviate the semantic compromise via multiview contrastive learning instead of introducing noise for graph augmentation.

\section{Methodology}
Details of the proposed method are given in this section, which is separated into subsections addressing five topics: an overview of the method, DP on graph edges, GCL with augmented graphs, the global parameter update process, and a summary. The notation necessary for understanding the proposed methodology is listed in Table \ref{tab:notation} for reference.

\begin{table}[t]
\setlength{\belowcaptionskip}{10pt}
\caption{Table of Notation}
\centering 
\footnotesize
\begin{tabular}{r c p{8cm}}
\toprule
\multicolumn{3}{c}{\underline{Notation}}\\
$\mathcal{G}$ & $\triangleq$ & A set of multiple graphs\\
$G_i$ & $\triangleq$ & The $i$-th graph, consisting of a node and an edge set\\
$\mathcal{V}_i$ & $\triangleq$ & The set of all nodes in the $i$-th graph\\
$\mathcal{E}_i$ & $\triangleq$ & The set of all edges in the $i$-th graph\\
$\mathbf{A}_i$ & $\triangleq$ & The adjacency matrix of the $i$-th graph\\
$\mathbf{X}_i$ & $\triangleq$ & The initial node features of the $i$-th graph\\
$p_i$ & $\triangleq$ & The total number of nodes in the $i$-th graph\\
$q$ & $\triangleq$ & The dimensionality of the initial node features\\
$\Delta f$ & $\triangleq$ & The sensitivity for DP\\
$Y$ & $\triangleq$ & The noise introduced for privacy protection, which is generated from a Laplacian distribution\\
$L(a, b)$ & $\triangleq$ & A Laplacian distribution with $a$ as the location parameter and $b$ as the scale parameter\\
$G_{i, 0}, G_{i, 1}$ & $\triangleq$ & Two augmented views of the $i$-th graph\\
$\mathbf{A}_{i, 0}, \mathbf{A}_{i, 1}$ & $\triangleq$ & The adjacency matrices of two augmented views of the $i$-th graph\\
\multicolumn{3}{c}{}\\
\multicolumn{3}{c}{\underline{Functions}}\\
$f$ & $\triangleq$ & A query function on graph edges\\
$M$ & $\triangleq$ & A privacy-aware query function on graph edges\\
$g$ & $\triangleq$ & A graph encoder\\
$p$ & $\triangleq$ & A classifier that predicts graph labels\\
\bottomrule
\end{tabular}
\label{tab:notation}
\end{table}

\subsection{Overview}

An overview of the proposed FGCL method is illustrated in Fig. \ref{fig:fl}, which also lists the five steps of the GCL process in a federated setting:

\begin{figure}[t]
	\centering
	\includegraphics[width=0.7\textwidth]{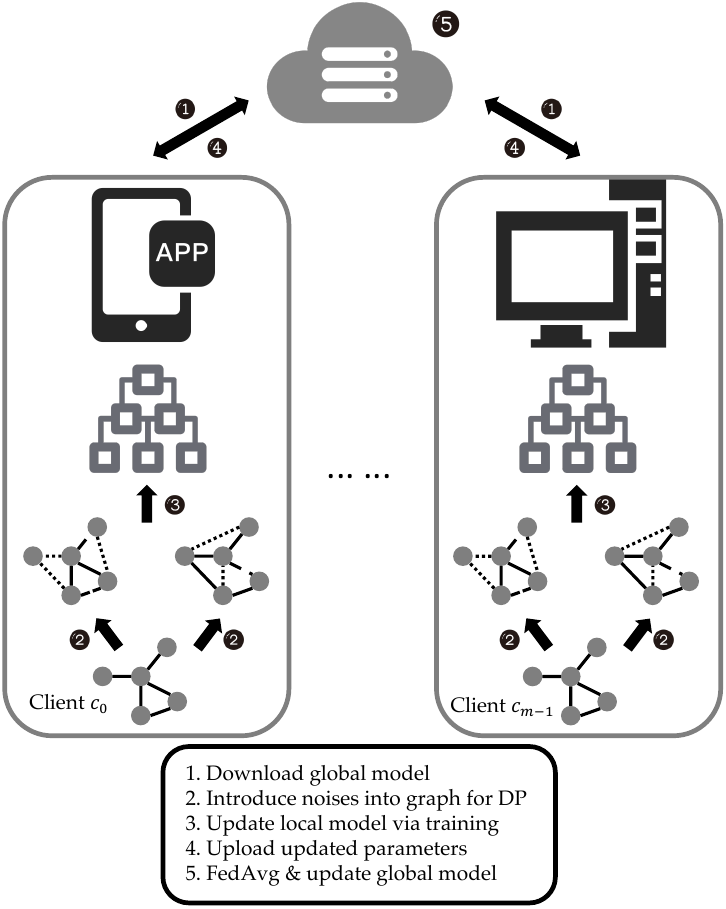}
\caption{The federated graph contrastive learning (FGCL) workflow with multiple clients.}
	\label{fig:fl}
\end{figure}

\noindent(1) \textbf{Download the global model.} The initial parameters of the graph encoder and the classifier are stored on the remote server. Clients must first download these parameters from the cloud before performing local training.

\noindent(2) \textbf{Introduce noise into the graph for DP.} The Laplacian mechanism is used to process each graph in the current training batch on every client twice to introduce noise into the graph proximity, thereby generating two augmented views of the original graph for GCL.

\noindent(3) \textbf{Update the local model via training.} Following the GCL training protocols, the model parameters downloaded from the server are updated.

\noindent(4) \textbf{Upload the updated parameters to the server.} The clients involved in the current training round upload their locally updated parameters to the server for a global model update.

\noindent(5) \textbf{FedAvg \& update the global model.} The server utilizes the FedAvg algorithm \cite{fedavg} to aggregate the updated local parameters and update the global parameters.
Specifically, the server collects the uploaded parameter updates and averages them to acquire the global updates.

Following the workflow above, we can train a global model on the server. Steps (2) and (3) are introduced in Section \ref{sec:method-dp} and Section \ref{sec:cl}, respectively, and step (5) is discussed in Section \ref{sec:method-summary}.

\subsection{DP on Graph Edges}
\label{sec:method-dp}
Some research works have adopted DP to introduce noise into the node or graph embeddings to protect privacy \cite{gw}. However, such strategies could be problematic when the initial node features are lacking. Instead, researchers must explore how to apply DP methods to graph structures (e.g., node privacy and edge privacy \cite{graph-privacy}). Here, we convert graph data into tabular data to enable the utilization of current DP methods, such as the Laplacian and Gaussian mechanisms. Without loss of generality, we consider only the Laplacian mechanism in this paper.

DP methods such as the Laplacian mechanism were originally designed for application to tabular data \cite{dp-challenge}. The key to adapting such methods to preserve graph structural privacy is to convert graph data into tabular data and to define the concept of `adjacent graphs' \cite{dp-survey}. In our method, a pair of node indices is treated as the key, and the connectivity between these two nodes is treated as the corresponding value to form a tabular dataset of key-value pairs. Regarding the definition of `adjacent graphs', two given graphs $G_1 = \{\mathcal{V}_1, \mathcal{E}_1\}$ and $G_2=\{\mathcal{V}_2, \mathcal{E}_2\}$ are considered `adjacent' if and only if $\mathcal{V}1=\mathcal{V}2$ and $||(\mathcal{E}_1\cup \mathcal{E}_2)-(\mathcal{E}_1\cap \mathcal{E}_2)||=1$ \cite{dp-survey}; an illustrative example is shown in Fig. \ref{fig:example}. Let $f(*)$ denote the query function on the tables generated from two `adjacent graphs'. To defend against differential attacks, we must introduce noise into the look-up results $M(G)=f(G)+Y$, where $Y$ denotes Laplacian random noise. Specifically, $Y\sim L(0, \frac{\Delta f}{\epsilon})$ satisfies $(\epsilon, 0)$-differential privacy, where $\Delta f=\max_{\{G_1, G_2\}}||f(G_1)-f(G_2)||=1$ is the sensitivity and $\epsilon$ is the privacy budget \cite{dp-proof}.

From the perspective of the graph structure, introducing noise into the entry values in the table generated from a graph is equivalent to perturbing the proximity of the graph. Let $\mathbf{A}$ denote the adjacency matrix of graph $G$; then, the perturbed adjacency matrix is
\begin{equation}
	\hat{\mathbf{A}}=\mathbf{A}+Y,
\end{equation}
where $Y\sim L(0, \frac{\Delta f}{\epsilon})$. Such privacy preservation results in a sacrifice in performance due to the introduction of noise into the graph. However, the process of introducing noise can be regarded as a type of graph augmentation, which is categorized as edge perturbation. The underlying prior for edge perturbation is semantic robustness against connectivity variations \cite{graphcl}. Given augmented views of the same graph, the aim of GCL is to maximize the agreement between these views via a contrastive loss function, such as InfoNCE \cite{infonce}, to force the graph encoder to acquire representations that are invariant with respect to the graph augmentations \cite{graphcl}. GCL can help to improve performance on various graph learning tasks, as empirically proven in previous works \cite{graphcl, gcc}. Therefore, we can leverage the advantages of GCL to alleviate the performance drop incurred with privacy preservation. However, it is worth noting that the entries in the adjacency matrix of a graph are binary and cannot take the decimal values that are derived from the original entries by adding noise. Therefore, in practice, we convert the original graph into a fully connected graph and store the protected proximity information as edge weights instead of using the adjacency matrix.

\subsection{GCL with Augmented Graphs}
\label{sec:cl}
A performance drop in graph learning is unavoidable when noise is introduced into a graph to perturb the graph structure \cite{performance-dropping}. However, various GCL techniques \cite{graphcl, dsgc, mvgrl, dgi, gca, gcc} have been developed that can facilitate the learning of invariant representations and the identification of critical structural semantics for augmented graphs, supported by rigorous theoretical and empirical analysis \cite{graphcl, gca}. Because the process of achieving graph-structural DP introduced in the previous section satisfies the definition of graph augmentation, we can utilize GCL to extract invariant representations from such noise-augmented graphs in order to mitigate the performance drop caused by the introduction of noise.

\begin{figure*}[htbp]
	\centering
	\includegraphics[width=\textwidth]{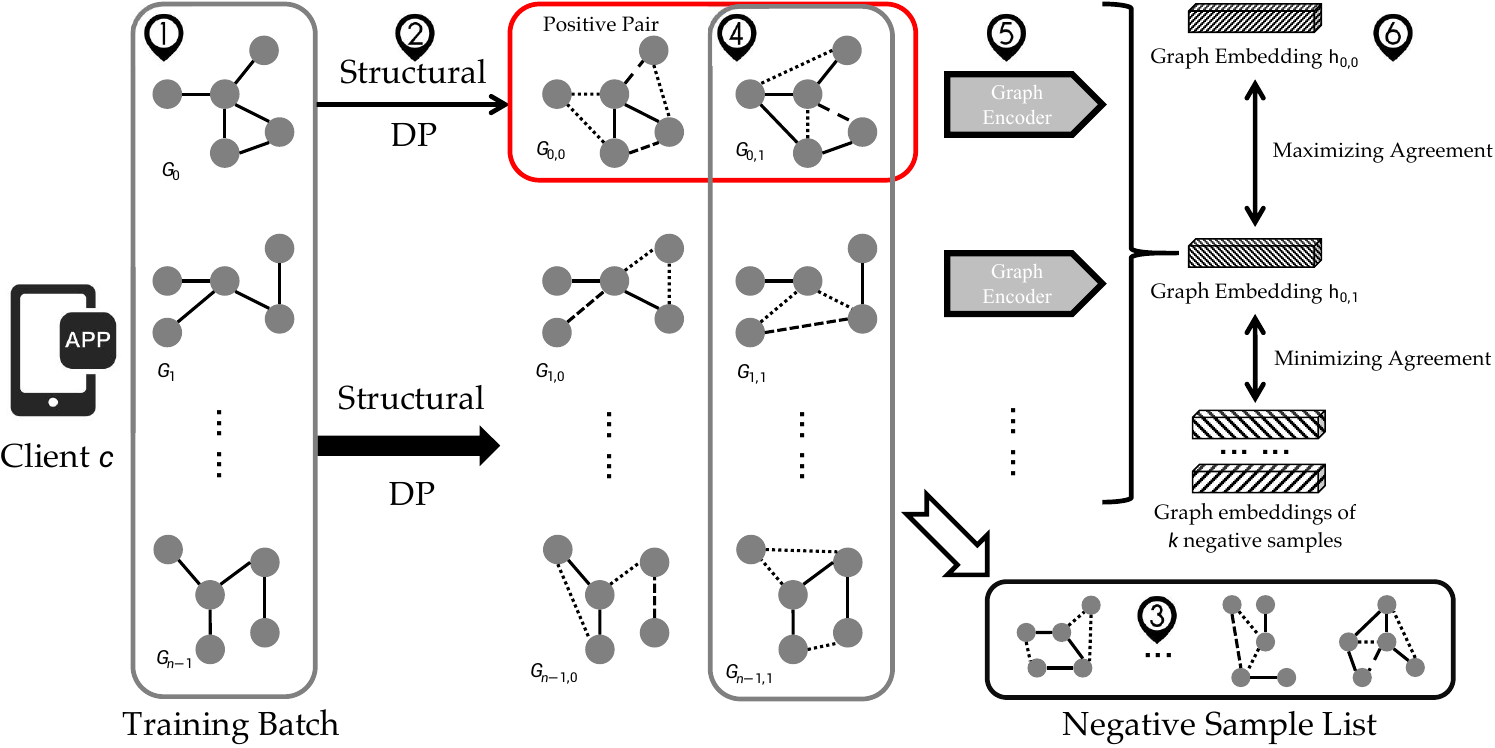}
\caption{An overview of GCL with augmented graphs on a single client, which consists of six steps: 1). Training batch partitioning. 2). Applying DP to the graph edges. 3). Maintaining a list to store negative samples. 4) Coupling of contrasting samples. 5) Graph encoding. 6) Learning objectives.}
	\label{fig:cl}
\end{figure*}

We first consider the scenario of conducting GCL with augmented graphs on a single client. As shown in Fig. \ref{fig:cl}, the whole process can be roughly broken down into six steps:

\noindent(1) \textbf{Training batch partitioning.} Each client possesses a local graph dataset $\mathcal{\mathcal{G}}=\{G_0, G_1, ..., G_n\}$. It is unlikely that the local model can be trained on all of the data simultaneously. In accordance with practical experience, we first need to determine the batch size $m$ to be used in partitioning the local dataset into training batches $\mathcal{\mathcal{G}}_b\subseteq\mathcal{\mathcal{G}}$, where $||\mathcal{\mathcal{G}}_b||=m$, at each client. Moreover, in many GCL methods \cite{graphcl, dsgc, gcc}, having negative contrasting samples with which to formulate negative pairs is mandatory to complete the GCL process. We can follow \cite{graphcl, dsgc} in sampling negative contrasting samples from the same training batch to which the positive sample belongs. However, if the model on a local device is trained in a streaming manner \cite{stream}, meaning that $m=1$, this method will no longer work. We address this limitation in step (3).

\noindent(2) \textbf{Applying DP to the graph edges.} Laplacian random noise is introduced in this step to perturb the graph proximity information to achieve DP. This process can be regarded as graph augmentation from the perspective of GCL. Graph $G_i$ in $\mathcal{\mathcal{G}}_b$ has adjacency matrix $\mathbf{A}_i$. We apply DP to the adjacency matrix twice to obtain two augmented views of the original graph, $G_{i, 0}$ and $G_{i, 1}$. The two augmented graphs have adjacency matrices $\mathbf{A}_{i, 0}=\mathbf{A}+Y_0$ and $\mathbf{A}_{i, 1}=\mathbf{A}+Y_1$, respectively, where $Y_0\sim L(0, \frac{\Delta f}{\epsilon_0})$ and $Y_1\sim L(0, \frac{\Delta f}{\epsilon_1})$. Note that the privacy budgets used to produce these two different augmented views could be different. Intuitively, GCL could potentially benefit more from maximizing the agreement between two more distinguishable views (e.g., produced with different privacy budgets). We will examine this intuition through comprehensive experiments.

\noindent(3) \textbf{Maintaining a list to store negative samples.} To address the limitation of traditional methods when $m=1$, as mentioned previously, we propose maintaining a list of a fixed size in which to store negative samples. Specifically, at the beginning of the training process, we initialize a list whose size is $N$. Then, we select the last $k$ instances, where $k<<N$, in the graph dataset and insert them into the list. These $k$ instances will serve as the negative samples for the target graph $G_t$ in the first round of training. Once a training round has finished, two perturbed views of the target graph will have been generated, namely, $G_{t, 0}$ and $G_{t, 1}$, and one of them will be inserted into the list. The noise applied to both views follows the same distribution, i.e., a Laplacian distribution. Theoretically, the edge weight matrices of both views should be the same. Therefore, either one of them can be inserted into the list. Once the number of elements in the list would be greater than $N$, the oldest element will be discarded. In all subsequent training rounds, all negative samples will be sampled from this list. Here, we specify the reason for adopting such a strategy. The current research literature describes many ways to acquire negative samples. For example, GCC \cite{gcc} samples negative instances from another dataset, CGC \cite{cgc} adaptively generates negative instances via a counterfactual mechanism, and GraphCL \cite{graphcl} samples negative instances from the training batch. However, the first two methods do not work in our setting. Sampling negative instances from other datasets is infeasible because the clients or servers in an FL system cannot arbitrarily access others. Generating negative samples locally could be problematic because most clients (e.g., smartphones and tablets) have limited computational resources. Therefore, the best way to acquire negative samples is by sampling them from the training batch to which the target belongs, following the proposed protocols in GraphCL \cite{graphcl}.

\noindent(4) \textbf{Coupling of contrasting samples.} Although some details of the acquisition of negative samples and positive samples have been introduced in the previous two steps, we give a formal description of the coupling of contrasting samples here. Given a target graph $G_i \in\mathcal{\mathcal{G}}_b$, we apply DP to $G_i$ to obtain two augmented views: $G_{i, 0}$ and $G_{i, 1}$. Then, we couple these two views $\{G_{i, 0}, G_{i, 1}\}$ as a \textit{positive pair} between which the agreement will be maximized. For negative samples, we follow the settings adopted in \cite{graphcl, dsgc}. Specifically, we sample $k$ negative graph instances $\{G^n_0, G^n_1, ..., G^n_{k-1}\}\subseteq\mathcal{\mathcal{G}}_b$, which should have different labels from that of the target graph. The sampled negative instances may be duplicated if the number of graphs whose labels are different from that of the target graph is less than $k$. One of the augmented views of each negative sample will be coupled with one of the augmented views of the target graph. Thus, without loss of generality, we form a set of negative pairs $\{\{G^n_{0, 1}, G_{i, 1}\}, \{G^n_{1, 1}, G_{i, 1}\}, ..., \{G^n_{k-1, 1}, G_{i, 1}\}\}$. The agreement between each negative pair will be minimized. Note that $G_{i, 0}$ and $G_{i, 1}$ are two augmented views of the same graph generated by the same DP mechanism; therefore, either $G_{i, 0}$ or $G_{i, 1}$ can be equivalently used to construct the negative pair set.

\begin{algorithm*}[htbp]
	\SetAlgoLined
	\caption{FGCL algorithm}
	\label{alg:overview}
	\SetKwFunction{READOUT}{READOUT}
	\KwIn{Parameters of the global model, $\theta^t$\;
		Training set for local client $i$, $\mathcal{G}_i=\{G_0, G_1, \cdots, G_n\}$\;
		Graph encoder, $f(\cdot, \cdot)$\;
		Readout function, \READOUT$(\cdot)$\;
		Similarity score function, $s(\cdot, \cdot)$\;
		Prediction function, $p(\cdot)$\;
		Temperature hyperparameter, $\tau$\;
		Weight control hyperparameter, $\gamma$\;
		The number of clients involved in a training round, $c$\;
		The total number of clients, $M$.}
	\KwOut{The updated global model.}
	\emph{Client:}\
	
	\emph{Download the global model}\;
	
	\For{each client $i\in \{0, 1, \cdots, i, \cdots, M-1\}$}{\emph{Update the local model}: $\theta_i\leftarrow\theta^t$;\
		
		\For{$G_j\in\mathcal{G}_i=\{G_0, G_1, \cdots, G_n\}$}{
			\emph{Apply DP on graph edges to obtain two perturbed views of $G_j$}: $G_{j, 0}, G_{j, 1}$;\
			
			\emph{Generate node embeddings}: $\mathbf{H}_{j, 0}=g(\mathbf{A}_{j, 0}, \mathbf{X}_j), \mathbf{H}_{j, 1}=g(\mathbf{A}_{j, 1}, \mathbf{X}_j)$;\
			\tcp{$\mathbf{A}$ is the adjacency matrix, and $\mathbf{X}$ is the feature matrix.}
			\emph{Generate graph embeddings}: $h_{j, 0}=$\READOUT$(\mathbf{H}_{j, 0}), h_{j, 1}=$\READOUT$(\mathbf{H}_{j, 1})$;\
			
			\emph{Contrastive learning objective}: $\mathcal{L}_c	=-\frac{1}{n}\sum_{i=0}^{n-1}\log\frac{e^{s(h_{i, 0}, h_{i, 1})/\tau}}{e^{s(h_{i, 0}, h_{i, 1})/\tau}+\sum_{t=0}^{k-1}e^{s(h^n_t, h_{i, 1})/\tau}}$;\
			
			\emph{Classification objective}: $\mathcal{L}_e=\frac{1}{2\cdot n}\sum_{i=0}^{n-1}(c(p(h_{j, 0}), labels_i)+c(p(h_{i, 1}), labels_i))$;\
			
			\emph{Overall training objective}: $\mathcal{L}=\gamma\cdot\mathcal{L}_c+\mathcal{L}_e$
			
			\emph{Backpropagate to update local parameters}: $\theta_i\leftarrow\theta_i^{t+1}$;\
		}
	}
	
	\emph{Server:}\
	
	\emph{Update the global model:} $\theta^{t+1}=\frac{c}{M}\sum\theta^{t+1}_i+\frac{M-c}{M}\sum\theta^t$;\
	
	\Return{$\theta^{t+1}$}
\end{algorithm*}

\noindent(5) \textbf{Graph encoding.} After obtaining a series of graphs, the next step is to encode these graphs to acquire high-quality graph embeddings. Various graph encoders have been proposed, such as GNN models. In this paper, we select three representative GNN models for study: GCN \cite{gcn}, GAT \cite{gat}, and GraphSAGE \cite{graphsage}. Let $g(\cdot, \cdot)$ denote the graph encoder, let $\mathbf{A}_i\in\mathcal{\mathbf{R}}^{p_i\times p_i}$ denote the proximity matrix of graph $G_i$,
and let $\mathbf{X}_i\in\mathcal{\mathbf{R}}^{p_i\times q}$ denote the node feature matrix of graph $G_i$, where $p_i$ is the number of nodes in $G_i$ and $q$ denotes the dimensionality of the initial node features. We can update the node embeddings by feeding the adjacency matrix and initial feature matrix into the graph encoder:
\begin{equation}
	\mathbf{H}_i=g(\mathbf{A}_i, \mathbf{X}_i)\in\mathcal{\mathbf{R}}^{p_i\times h},
\end{equation}
where $h$ denotes the hidden dimension. Then, a readout function will be applied to summarize the node embeddings to obtain the graph embedding:
\begin{equation}
	h_i=READOUT(\mathbf{H}_i)\in\mathcal{\mathbf{R}}^{1\times h}.
\end{equation}
There are many choices for the readout function, and the chosen function could vary among different downstream tasks. To obtain the embeddings of selected positive pairs and negative pairs, we apply graph encoding to these graphs to obtain positive embedding pairs $\{h_{i, 0}, h_{i, 1}\}$ and negative embedding pairs $\{\{h^n_{0, 1}, h_{i, 1}\}, \{h^n_{1, 1}, h_{i, 1}\}, ..., \{h^n_{k-1, 1}, h_{i, 1}\}\}$.

\noindent(6) \textbf{Learning objectives.} The objective of GCL can be summarized as maximizing the agreement between positive pairs and minimizing the agreement between negative pairs. We choose InfoNCE \cite{infonce}, which is widely used in many works \cite{gcc, dsgc}, to serve as the objective function for GCL:
\begin{equation}
	\mathcal{L}_c	=-\frac{1}{n}\sum_{i=0}^{n-1}\log\frac{e^{s(h_{i, 0}, h_{i, 1})/\tau}}{e^{s(h_{i, 0}, h_{i, 1})/\tau}+\sum_{t=0}^{k-1}e^{s(h^n_t, h_{i, 1})/\tau}},
\end{equation}
where $s(\cdot, \cdot)$ is a function measuring the similarity of two embeddings, such as the cosine similarity, and $\tau$ serves as the temperature hyperparameter to adjust the contrastive loss. Moreover, for scenarios in which the training data have labels, we can adopt the cross-entropy function $c(\cdot, \cdot)$ to introduce supervision signals into the training process. First, however, we need a classifier $p(\cdot)$ to predict labels in accordance with the obtained graph embeddings:
\begin{equation}
	res_{i, 0} = p(h_{i, 0}),
\end{equation}
\begin{equation}
	res_{i, 1} = p(h_{i, 1}).
\end{equation}
Then, we can apply the cross-entropy function:
\begin{equation}
	\mathcal{L}_e=\frac{1}{2\cdot n}\sum_{i=0}^{n-1}(c(res_{i, 0}, labels_i)+c(res_{i, 1}, labels_i)).
\end{equation}
For client $c$ and training batch $\mathcal{\mathcal{G}}_b$, we have the following training objective:
\begin{equation}
	\mathcal{L}=\gamma\cdot\mathcal{L}_c+\mathcal{L}_e,
\end{equation}
where $\gamma$ is a hyperparameter that controls the weight of the contrastive learning loss in the overall objective.

Each client in the proposed FL system follows the same training protocol described above to conduct GCL locally. By leveraging the advantages of GCL, we expect to distil critical structural semantics from the augmented graphs in order to obtain graph representations that will remain invariant after the introduction of noise, thereby acquiring high-quality graph embeddings for the downstream task.

\subsection{Global Parameter Update}
\label{sec:method-summary}

When the training phase has finished on each client, the updated local parameters will be uploaded to the remote server. Then, the remote server will execute an FL algorithm to aggregate these local updates and update the global model. For simplicity, we consider the most widely used algorithm for this purpose, FedAvg \cite{fedavg}, in this paper.

Specifically, the server collects the uploaded parameter updates and averages them to acquire the global updates. Let $\theta^{t+1}$ and $\theta^t$ denote the global parameters at times $t+1$ and $t$, respectively, and let $\theta^{t+1}_i$ denote the updated parameters of the local model stored on client $c_i$. Suppose that there are $M$ clients in total and that $c$ clients will be sampled during this update, denoted by the set $\mathcal{\mathbf{S}}_c$, to participate in training in each round. Thus, the process of FedAvg can be formulated as follows:
\begin{equation}
\theta^{t+1}=\frac{c}{M}\sum_{c_i\in\mathcal{\mathbf{S}}_c}\theta^{t+1}_i+\frac{M-c}{M}\sum_{c_i\not\in\mathcal{\mathbf{S}}_c}\theta^t
\end{equation}

\subsection{Summary}
After the whole training process is completed, the trained model can be used to conduct inference. Note that there is no difference between the training and inference processes. Moreover, we summarize the whole training procedure in Algorithm \ref{alg:overview} to better illustrate the proposed methodology.

\section{Experiments}
Detailed experimental settings are listed for reproducibility at the beginning of this section, followed by an introduction to the datasets and base models we adopted in the experiments. Then, we report comprehensive experiments and give the corresponding analysis, providing a detailed view of the proposed FGCL method.

\begin{table}[t]
	\centering
	\setlength{\belowcaptionskip}{10pt}
	\caption{Definitions of the hyperparameters}
	\label{tab:hyper}
        \footnotesize
	\begin{tabular}{r c p{8cm}}
		\toprule
		\multicolumn{3}{c}{\underline{Hyperparameters}}\\
  \multicolumn{3}{c}{}\\
            h & $\triangleq$  & The hidden dimension size.    \\
            m  & $\triangleq$      & The size of a training batch.     \\
		N   & $\triangleq$     & The size of the negative sample stack.  \\
		k   & $\triangleq$     & The number of negative samples used in each GCL training round.         \\
		$\epsilon$ & $\triangleq$ & The privacy budget.             \\
		$\gamma$  & $\triangleq$  & The hyperparameter that controls the weight of GCL in the overall training objective. \\
		$\tau$  & $\triangleq$    & The temperature parameter in the contrastive learning objective. \\ \bottomrule
	\end{tabular}
\end{table}

\subsection{Experimental Settings}
The proposed FGCL method focuses on federated graph learning. To implement the proposed method, a widely used toolkit named \textit{FedGraphNN} \footnote{https://github.com/FedML-AI/FedGraphNN} is adopted. It provides various APIs and exemplars to help build federated graph learning models quickly. FGCL shares the same training protocols as \textit{FedGraphNN} as well as some common hyperparameters, including the hidden dimension size, the number of GNN layers, and the learning rate. Notably, only the graph classification task is used for measurement in our experiments. The detailed settings can be found in \cite{fedml, fedgraphnn}.

Nevertheless, some hyperparameters are unique to the GCL module and therefore need to be specified for reproducibility. The definitions and values of these hyperparameters can be found in Table \ref{tab:hyper}. 
Please refer to the source code\footnote{\url{https://www.dropbox.com/sh/ql9m1d6lltygpgl/AADwD3VtTE_9S2cmypa7wR34a?dl=0}} for more details.

\subsection{Datasets and Baselines}
In this subsection, we give a brief introduction to the datasets and baselines used in this work.
\subsubsection{Datasets}
We chose several widely used and well-known graph datasets to conduct our experiments, as follows:
\begin{itemize}
\item \textbf{SIDER} \footnote{http://sideeffects.embl.de/} \cite{sider} contains information about medicines and their observed adverse drug reactions. This information includes drug classification.
\item \textbf{BACE} \cite{bace} provides qualitative binding results (binary labels) for a set of inhibitors of human beta-secretase 1.
\item \textbf{ClinTox} \cite{clintox} compares drugs approved by the FDA with drugs that have failed clinical trials for toxicity reasons. It assigns a binary label to each drug molecule to indicate whether it exhibited toxicity.
\item \textbf{BBBP} \cite{bbbp} is designed for the modeling and prediction of barrier permeability. It allocates binary labels to drug molecules to indicate whether they are able to penetrate the blood--brain barrier.
\item \textbf{Tox21} \footnote{https://tripod.nih.gov/tox21/challenge/} contains graphs of chemical compounds. Each compound has 12 labels to reflect outcomes in 12 different toxicological experiments.
\end{itemize}

The statistics of the datasets mentioned are summarized in Table \ref{tab:statistics}.

\begin{table*}[htbp]
	\centering
	\renewcommand{\arraystretch}{1.4}
        \setlength{\belowcaptionskip}{10pt}
	\caption{Statistics of the datasets}
	\label{tab:statistics}
	\footnotesize
		\begin{tabular}{c|ccccc}
			\toprule
			Dataset & \# Graphs & Avg. \# Nodes & Avg. \# Edges & Avg. Degree & \# Classes \\ \midrule
			SIDER   & 1427      & 33.64         & 35.36         & 2.10           & 27         \\
			BACE    & 1513      & 34.12         & 36.89         & 2.16           & 2          \\
			ClinTox & 1478      & 26.13         & 27.86         & 2.13           & 2          \\
			BBBP    & 2039      & 24.05         & 25.94         & 2.16           & 2          \\
			Tox21   & 7831      & 18.51         & 25.94         & 2.80           & 12         \\ \bottomrule
	\end{tabular}
\end{table*}

\subsubsection{Baselines}
Since the performance of federated graph learning may vary with different graph encoders, we adopted several widely used GNN models to serve as graph encoders to comprehensively examine the proposed method. Moreover, we leveraged the advantages of the toolkit named \textit{PyTorch Geometric} \footnote{https://www.pyg.org/} to efficiently implement high-quality GNN models.

Note that as mentioned in Section \ref{sec:method-dp}, the protected proximity information is stored as edge weights. Hence, we selected only GNNs in \textit{PyTorch Geometric} that are capable of handling edge weights, as follows:
\begin{itemize}
\item \textbf{GCN} \cite{gcn} is one of the first proposed GNN models, implementing a first-order approximation of spectral graph convolutions.
\item \textbf{kGNNs} \cite{kgnns} considers higher-order graph structures at multiple scales, thus overcoming the shortcomings of conventional GNNs.
\item \textbf{TAG} \cite{tag} is a novel graph convolutional network defined in the vertex domain, thereby alleviating the computational complexity of spectral graph convolutional neural networks.
\item \textbf{LGCN} \cite{lightgcn} is a simplified and lightweight version of GCN that includes only the most essential components.
\end{itemize}

\subsection{Experimental Results}
In this subsection, we present the experimental results, and a detailed and comprehensive analysis of the results is also given to illustrate the properties of the proposed method and justify FGCL's effectiveness.

\begin{table*}[htbp]
	\centering
	\renewcommand{\arraystretch}{1.6}
	\setlength{\belowcaptionskip}{2pt}
	\caption{Results of comparative experiments (ROC-AUC).}
	\label{tab:comparison}
	\resizebox{\textwidth}{!}{
	\begin{tabular}{c|c|ccccc}
		\hline
		\multirow{2}{*}{Base Models}                        & \multirow{2}{*}{Settings} & \multicolumn{5}{c}{Datasets}                                                                                                                                                                       \\ \cline{3-7} 
		&                           & SIDER                                & BACE                                 & ClinTox                               & BBBP                                 & Tox21                                 \\ \hline
		\multirow{10}{*}{GCN}                           & Centralized               & 0.6637                               & 0.8154                               & 0.9227                                & 0.8214                               & 0.7990                                \\
		& Federated                 & 0.6055                               & 0.6373                               & 0.8309                                & 0.6576                               & 0.5338                                \\ \cline{2-7} 
		& Fed+Noise / 100           & 0.6050                               & 0.6162                               & 0.8003                                & 0.6632                               & 0.5317                                \\
		& Fed+Noise / 10            & 0.5884                               & 0.6245                               & 0.7598                                & 0.6577                               & 0.5303                                \\
		& Fed+Noise / 1             & 0.5636                               & 0.6246                               & 0.7061                                & 0.6185                               & 0.5327                                \\
		& Fed+Noise / 0.1           & 0.5374                               & 0.6666                               & 0.7091                                & 0.6036                               & 0.5323                                \\ \cline{2-7} 
		& FGCL+Noise / 100          & 0.6285 (+3.88\%)                     & 0.6697 (+8.68\%)                     & 0.6740 (-15.78\%)                     & 0.7035 (+6.08\%)                     & 0.6675 (+25.54\%)                     \\
		& FGCL+Noise / 10           & 0.6334 (+7.65\%)                     & 0.6633 (+6.21\%)                     & 0.6999 (-7.88\%)                      & 0.6915 (+5.14\%)                     & 0.6697 (+24.17\%)                     \\
		& FGCL+Noise / 1            & 0.6062 (+7.56\%)                     & 0.6665 (+6.71\%)                     & 0.7917 (+12.12\%)                     & 0.6980 (+12.85\%)                    & 0.6470 (+21.47\%)                     \\
		& FGCL+Noise / 0.1          & 0.5549 (+3.26\%)                     & 0.6858 (+2.88\%)                     & 0.8378 (+18.15\%)                     & 0.7044 (+16.70\%)                    & 0.5961 (+11.99\%)                     \\ \hline
		\multirow{10}{*}{kGNNs}                         & Centralized               & 0.6785                               & 0.8912                               & 0.9277                                & 0.8541                               & 0.7687                                \\
		& Federated                 & 0.6033                               & 0.6131                               & 0.8787                                & 0.6371                               & 0.5310                                \\ \cline{2-7} 
		& Fed+Noise / 100           & 0.6035                               & 0.6296                               & 0.8108                                & 0.6526                               & 0.5499                                \\
		& Fed+Noise / 10            & 0.5817                               & 0.647                                & 0.7502                                & 0.6097                               & 0.5349                                \\
		& Fed+Noise / 1             & 0.5663                               & 0.6206                               & 0.8160                                & 0.5906                               & 0.5380                                \\
		& Fed+Noise / 0.1           & 0.5345                               & 0.6291                               & 0.7421                                & 0.5920                               & 0.5334                                \\ \cline{2-7} 
		& FGCL+Noise / 100          & 0.6140 (+1.74\%)                     & 0.6826 (+8.42\%)                     & 0.7037 (-13.21\%)                     & 0.7314 (+12.07\%)                    & 0.6720 (+22.20\%)                     \\
		& FGCL+Noise / 10           & 0.6223 (+6.98\%)                     & 0.6803 (5.15\%)                      & 0.7034 (-6.24\%)                      & 0.7264 (+19.14\%)                    & 0.6615 (+23.67\%)                     \\
		& FGCL+Noise / 1            & 0.6025 (+6.39\%)                     & 0.6789 (+9.39\%)                     & 0.7093 (-13.01\%)                     & 0.6817 (+15.42\%)                    & 0.6535 (+21.47\%)                     \\
		& FGCL+Noise / 0.1          & 0.5287 (-1.09\%)                     & 0.6433 (+2.26\%)                     & 0.8433 (+17.86\%)                     & 0.6356 (+7.36\%)                     & 0.6352 (+7.36\%)                      \\ \hline
		\multirow{10}{*}{TAG}                           & Centralized               & 0.6276                               & 0.7397                               & 0.9392                                & 0.7642                               & 0.6052                                \\
		& Federated                 & 0.5552                               & 0.6286                               & 0.8865                                & 0.6535                               & 0.5360                                \\ \cline{2-7} 
		& Fed+Noise / 100           & 0.5509                               & 0.6539                               & 0.8812                                & 0.6208                               & 0.5335                                \\
		& Fed+Noise / 10            & 0.5482                               & 0.6324                               & 0.7529                                & 0.5808                               & 0.5416                                \\
		& Fed+Noise / 1             & 0.5543                               & 0.6297                               & 0.7915                                & 0.6370                               & 0.5405                                \\
		& Fed+Noise / 0.1           & 0.5551                               & 0.5839                               & 0.7700                                & 0.6173                               & 0.5253                                \\ \cline{2-7} 
		& FGCL+Noise / 100          & 0.6174 (+12.07\%)                    & 0.7091 (+8.44\%)                     & 0.9034 (+2.52\%)                      & 0.7005 (+12.84\%)                    & 0.5592 (+4.82\%)                      \\
		& FGCL+Noise / 10           & 0.6369 (+16.18\%)                    & 0.6744 (+6.64\%)                     & 0.7923 (+5.23\%)                      & 0.7044 (+21.28\%)                    & 0.5463 (+0.87\%)                      \\
		& FGCL+Noise / 1            & 0.5619 (+1.37\%)                     & 0.6977 (+10.80\%)                    & 0.7387 (-6.67\%)                      & 0.6755 (+6.04\%)                     & 0.5342 (-1.17\%)                      \\
		& FGCL+Noise / 0.1          & 0.5721 (+3.06\%)                     & 0.6720 (+15.09\%)                    & 0.5978 (-22.36\%)                     & 0.6726 (+8.96\%)                     & 0.5359 (+2.02\%)                      \\ \hline
		\multicolumn{1}{l|}{\multirow{10}{*}{LGCN}} & Centralized               & 0.6866                               & 0.8385                               & 0.9269                                & 0.7829                               & 0.7734                                \\
		\multicolumn{1}{l|}{}                           & Federated                 & 0.5933                               & 0.6741                               & 0.8098                                & 0.6930                               & 0.5379                                \\ \cline{2-7} 
		\multicolumn{1}{l|}{}                           & Fed+Noise / 100           & 0.5916                               & 0.6863                               & 0.7785                                & 0.6632                               & 0.5350                                \\
		\multicolumn{1}{l|}{}                           & Fed+Noise / 10            & 0.5942                               & 0.6619                               & 0.6882                                & 0.6832                               & 0.5324                                \\
		\multicolumn{1}{l|}{}                           & Fed+Noise / 1             & 0.5747                               & 0.6157                               & 0.7938                                & 0.5826                               & 0.5358                                \\
		\multicolumn{1}{l|}{}                           & Fed+Noise / 0.1           & 0.5610                               & 0.5952                               & 0.6719                                & 0.6017                               & 0.5385                                \\ \cline{2-7} 
		\multicolumn{1}{l|}{}                           & FGCL+Noise / 100          & \multicolumn{1}{l}{0.6331 (+7.01\%)} & \multicolumn{1}{l}{0.6898 (+0.51\%)} & \multicolumn{1}{l}{0.8716 (+11.96\%)} & \multicolumn{1}{l}{0.7229 (+9.00\%)} & \multicolumn{1}{l}{0.6324 (+18.21\%)} \\
		\multicolumn{1}{l|}{}                           & FGCL+Noise / 10           & 0.6335 (+10.23\%)                    & 0.7093 (+7.16\%)                     & 0.7291 (+5.94\%)                      & 0.7138 (+4.48\%)                     & 0.6396 (+20.14\%)                     \\
		\multicolumn{1}{l|}{}                           & FGCL+Noise / 1            & 0.6016 (+4.68\%)                     & 0.6870 (+11.58\%)                    & 0.8027 (+1.12\%)                      & 0.7450 (27.88\%)                     & 0.6474 (+20.22\%)                     \\
		\multicolumn{1}{l|}{}                           & FGCL+Noise / 0.1          & 0.5678 (+1.21\%)                     & 0.6826 (+14.68\%)                    & 0.7882 (+17.31\%)                     & 0.6547 (+8.81\%)                     & 0.5823 (+8.13\%)                      \\ \hline
	\end{tabular}}
\end{table*}

\subsubsection{How much performance degradation does graph-structural DP cause in federated graph learning methods?}
Each base model is represented by three rows in Table \ref{tab:comparison}, which give the experimental results in centralized and federated settings, in federated settings with DP, and in the proposed FGCL settings. Here, we first examine the first two rows in this section.

First, we notice that the centralized setting outperforms the federated setting on all datasets for any base model. This phenomenon has been verified in many research works \cite{fedml, fedgraphnn, dp-privrec}. The reason is that the federated learning protocol distributes the training data to different devices and updates the global model based on the gradients or parameters passed back from the local devices, which makes the model training more biased.

Next, we consider the results of introducing noise generated from a Laplacian distribution to achieve DP on the graph edges. Specifically, we applied privacy budgets of $\{0.1, 1, 10, 100\}$ to generate noise via the DP mechanism. The corresponding results are shown in the second row. In most cases, the ROC-AUC scores in a federated setting with DP privacy preservation are lower than those in the pure federated setting. It is reasonable that the introduced noise protects privacy, but it simultaneously undermines the semantics in the graph, resulting in a performance decrease. Moreover, different privacy budgets are tested in these experiments. Theoretically, a greater privacy budget means more minor noise, indicating that a federated setting with a greater privacy budget should perform better than a federated setting with a lesser privacy budget. However, the experimental results do not strictly follow this pattern. This is because the noise is randomly generated. Consequently, this pattern is not expected to manifest until sufficient rounds of experiments are completed.

In summary, FL addresses some of the limitations of centralized training protocols, but it sacrifices performance. Moreover, introducing privacy-preserving mechanisms such as DP into FL further decreases performance. Hence, it is critical for the research community to explore solutions to alleviate this performance drop.

\subsubsection{Can GCL help alleviate the performance sacrifice caused by graph-structural DP?}
As mentioned in the previous subsection, introducing noise via the DP mechanism to protect privacy will decrease the performance of federated graph learning. In this paper, we propose leveraging the advantages of GCL to achieve semantic robustness against noise by contrasting perturbed graph views. The experimental results of the proposed method are also listed in Table \ref{tab:comparison}, in the third row for each base model. Overall, although the proposed FGCL setting is not competitive compared to the centralized setting, it outperforms the federated setting in some cases and is superior to the federated setting with DP privacy preservation in most cases.
The improvement is generally between 3\% and 15\%, but we notice that in some cases, either there is no improvement or the improvement is exceptionally high, sometimes more than 20\%. The cases of nonimprovement mainly occur in the experiments on the ClinTox dataset. We observe that the performances on the ClinTox dataset in the centralized setting, the federated setting, and the federated setting with DP noise are very high, much better than those on other datasets. Therefore, the room for improvement in performance on this dataset is less than that on the others. The findings suggest that our proposed FGCL method may not work well when the room for improvement is limited.
Regarding the cases of exceptionally high improvement, these appear in the experimental results on the Tox21 dataset. We note that on Tox21, the performance decrease after implementing the federated setting (with or without the DP mechanism) is the most severe, at approximately -35\%. This significant performance gap provides considerable room for improvement. Accordingly, the findings indicate that FGCL is more powerful when the performance drops dramatically with the implementation of federated learning and the DP mechanism.
Moreover, a particular case is worthy of note: the base model TAG. The improvement achieved through FGCL with TAG as the base model on the Tox21 dataset is minor. This is because of the nature of TAG. In the centralized setting, TAG has worse performance than the others, indicating that TAG is not as delicate as the others. We also find that the overall performance of FGCL with TAG as the base model is the worst among all base models, as will be illustrated in the next subsection. Hence, we conclude that choosing an appropriate base model is one of the keys to fully leveraging the advantages of the proposed FGCL method.

\begin{figure}[]
	\centering
	\subfigure[SIDER]{
		\includegraphics[width=3cm]{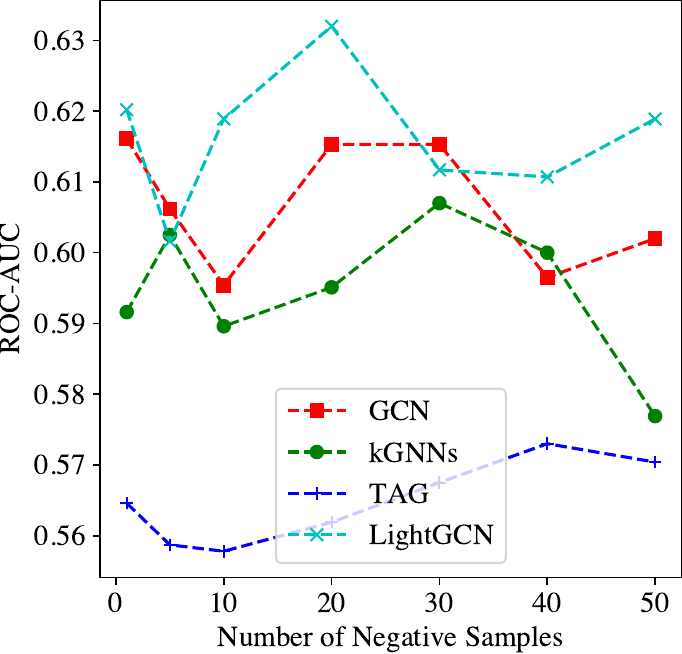}
	}
	\quad
	\subfigure[BACE]{
		\includegraphics[width=3cm]{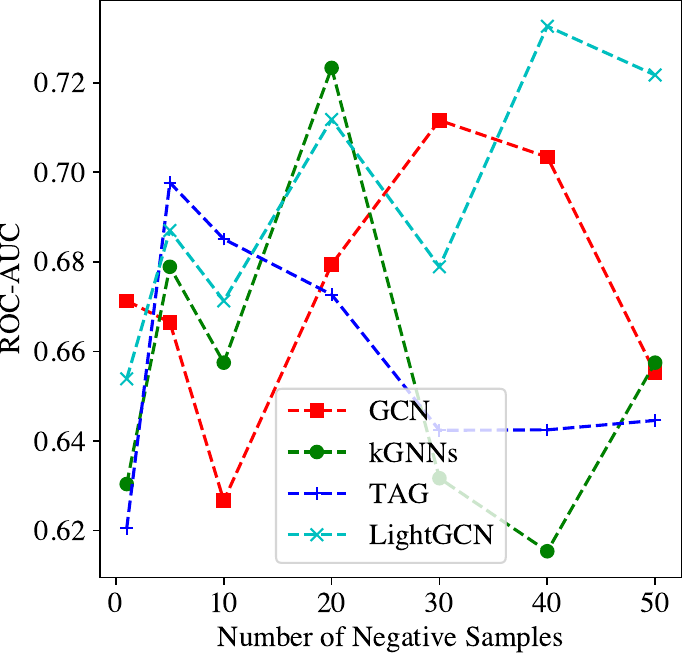}
	}
	\quad
	\subfigure[ClinTox]{
		\includegraphics[width=3cm]{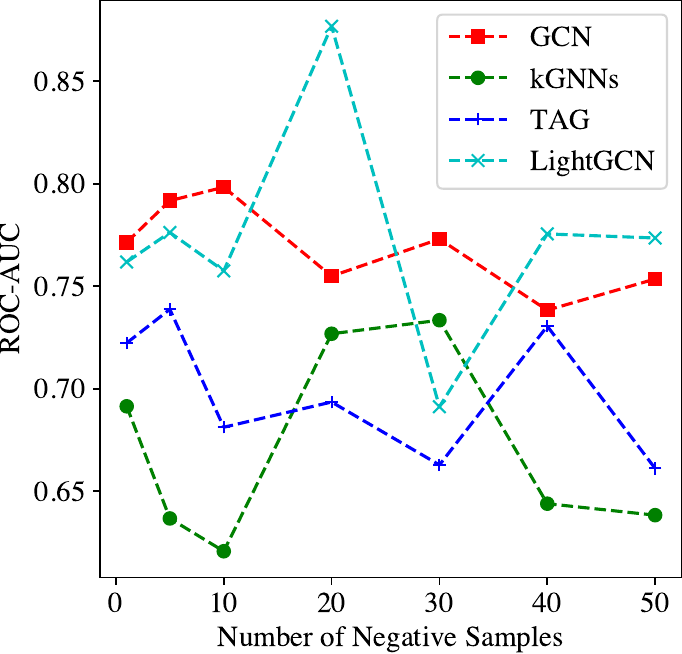}
	}
	\quad
	\subfigure[BBBP]{
		\includegraphics[width=3cm]{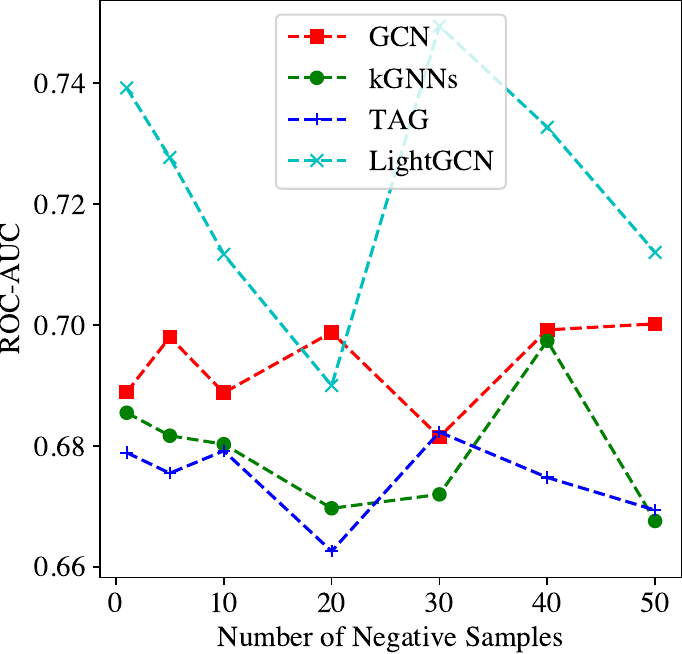}
	}
	\quad
	\subfigure[Tox21]{
		\includegraphics[width=3cm]{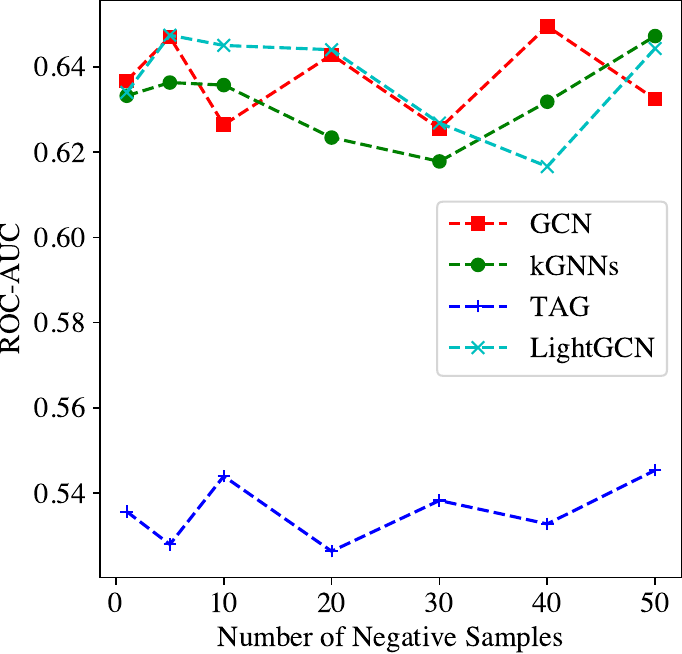}
	}
\caption{The number of negative samples is selected from the set [1, 5, 10, 20, 30, 40, 50]. The figures above show the performance of FGCL in combination with the different base models on all datasets with different numbers of negative samples.}
	\label{fig:negatives}
\end{figure}

\subsubsection{Study of the hyperparameters for FGCL}
Three hyperparameter experiments are reported in this section to comprehensively reveal the detailed characteristics and properties of the proposed FGCL method. The first hyperparameter investigated is $k$, the number of negative samples in the GCL module. Negative samples have been verified to play a critical role in the model training process \cite{negative}. Here, we fix all hyperparameters other than $k$ to investigate its impact. The candidate values of $k$ are $[1, 5, 10, 20, 30, 40 ,50]$, and the experimental results are illustrated in Figure \ref{fig:negatives}. The first observation is that selecting a suitable number of negative samples can further enhance the performance of FGCL and achieve greater improvement. However, there is no obvious pattern that would indicate how to select $k$, suggesting that this task is highly task- or dataset-specific. For instance, the BACE dataset requires FGCL to have a relatively large $k$ value for better performance, but a relatively small $k$ value is sufficient for the Tox21 dataset. In summary, the hyperparameter $k$ is highly task- or dataset-specific and should be carefully selected, and for computational efficiency, selecting a smaller $k$ when possible would be better.

\begin{figure*}[htbp]
	\centering
	\subfigure[SIDER]{
		\includegraphics[width=3.5cm]{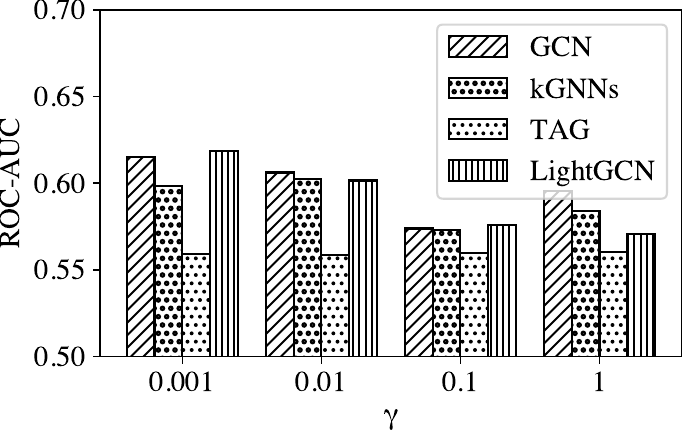}
	}
	\quad
	\subfigure[BACE]{
		\includegraphics[width=3.5cm]{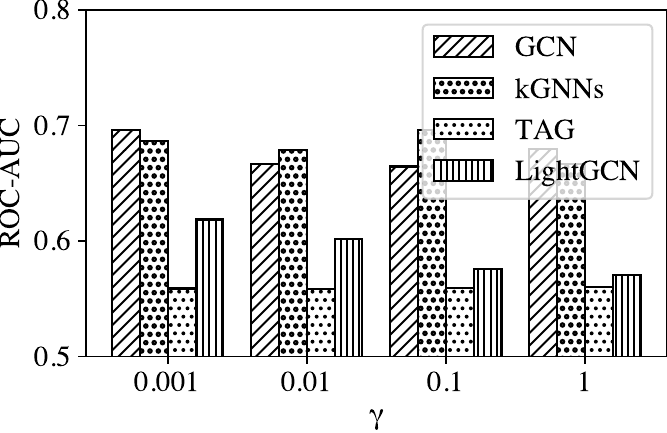}
	}
	\quad
	\subfigure[ClinTox]{
		\includegraphics[width=3.5cm]{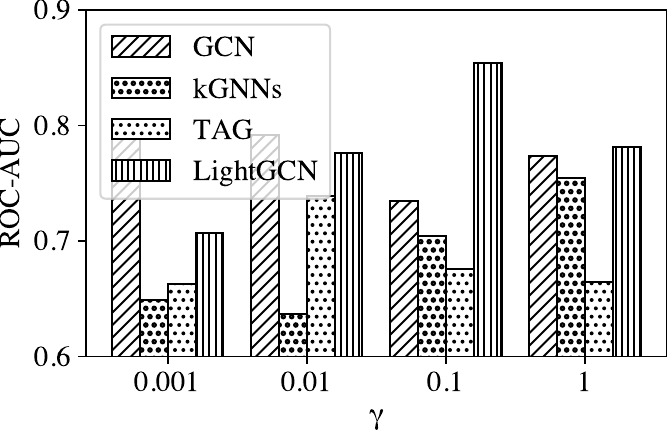}
	}
	\quad
	\subfigure[BBBP]{
		\includegraphics[width=3.5cm]{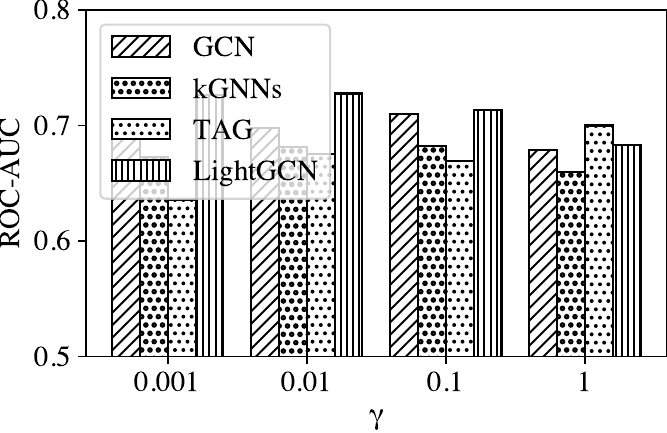}
	}
	\quad
	\subfigure[Tox21]{
		\includegraphics[width=3.5cm]{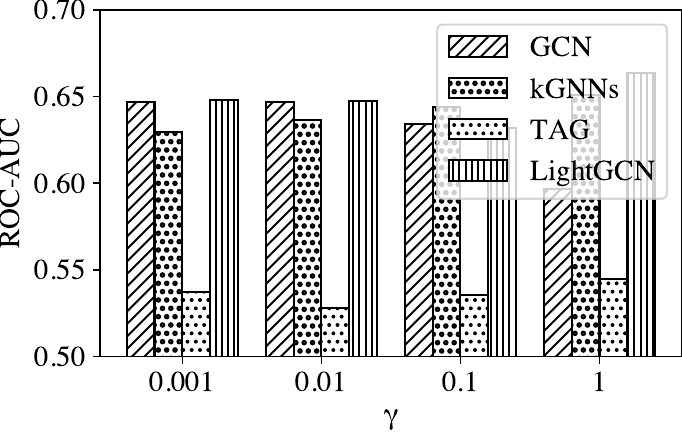}
	}
\caption{The value of $\gamma$ is selected from the value set [0.001, 0.01, 0.1, 1]. The figures show the performance of FGCL in combination with different base models on all datasets with different $\gamma$ values.}
	\label{fig:gamma}
\end{figure*}

The second hyperparameter experiment concerns $\gamma$, the parameter controlling the weight of the contrastive learning objective. The overall training objective of FGCL consists of two components: the graph classification loss and the contrastive learning loss. Between the two, graph classification is the primary task. $\gamma$ plays the role of balancing these two objectives. We select the value of $\gamma$ from the set $[0.001, 0.01, 0.1, 1]$, and the corresponding experimental results are shown in Figure \ref{fig:gamma}. According to the results, the impact of $\gamma$ on the performance is not as significant as that of $k$. Nevertheless, we observe that FGCL with $\gamma<1$ can achieve slightly higher ROC-AUC scores. When $\gamma\geq1$, the contrastive learning objective undermines the importance of the primary graph classification task in the overall objective, resulting in suboptimal performance on this task. Therefore, we recommend adopting a relatively small $\gamma$ value to achieve better results in practice.

\begin{figure*}[htbp]
	\centering
	\subfigure[GCN-SIDER]{
		\includegraphics[width=3cm]{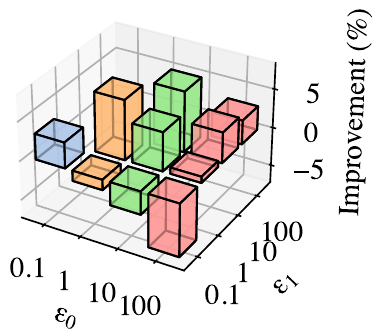}
	}
        \subfigure[kGNNs-SIDER]{
		\includegraphics[width=3cm]{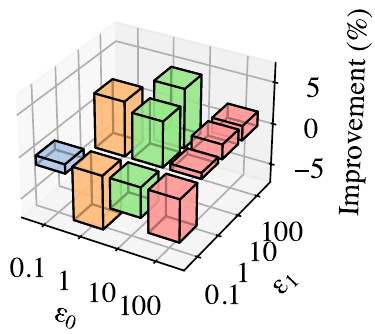}
	}
        \subfigure[TAG-SIDER]{
		\includegraphics[width=3cm]{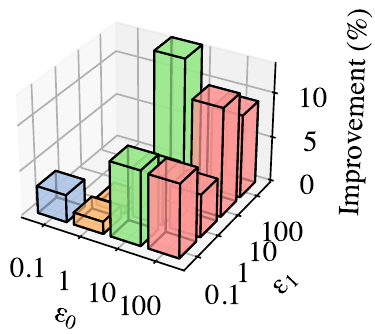}
	}
        \subfigure[LGCN-SIDER]{
		\includegraphics[width=3cm]{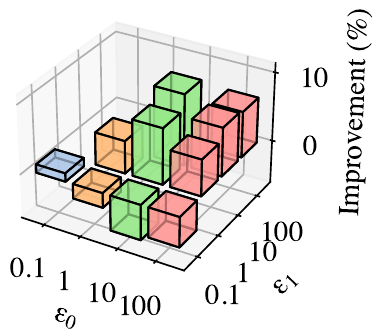}
	}
 
	\subfigure[GCN-BACE]{
		\includegraphics[width=3cm]{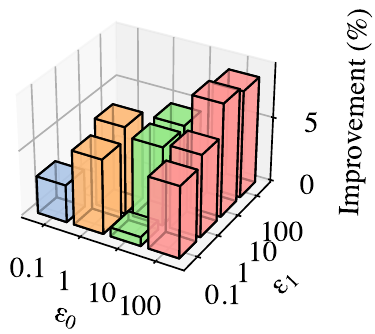}
	}
        \subfigure[kGNNs-BACE]{
		\includegraphics[width=3cm]{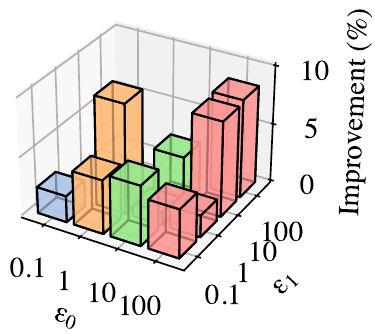}
	}
        \subfigure[TAG-BACE]{
		\includegraphics[width=3cm]{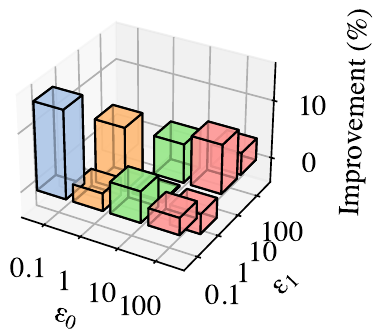}
	}
        \subfigure[LGCN-BACE]{
		\includegraphics[width=3cm]{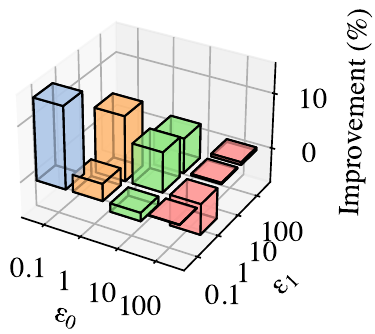}
	}
 
	\subfigure[GCN-ClinTox]{
		\includegraphics[width=3cm]{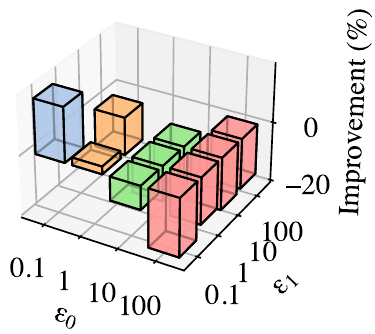}
	}
        \subfigure[kGNNs-ClinTox]{
		\includegraphics[width=3cm]{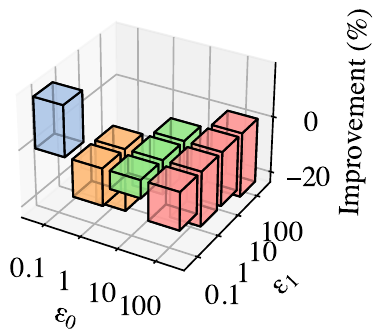}
	}
        \subfigure[TAG-ClinTox]{
		\includegraphics[width=3cm]{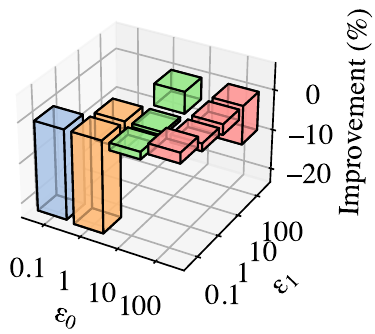}
	}
        \subfigure[LGCN-ClinTox]{
		\includegraphics[width=3cm]{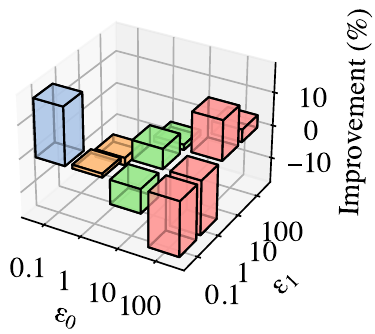}
	}
 
        \subfigure[LGCN-BBBP]{
		\includegraphics[width=3cm]{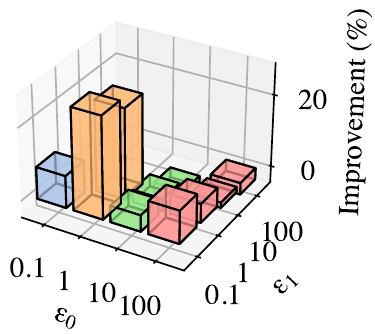}
	}
	\subfigure[GCN-BBBP]{
		\includegraphics[width=3cm]{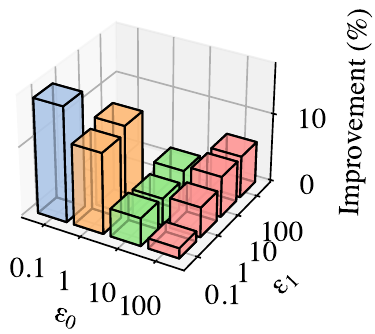}
	}
        \subfigure[kGNNs-BBBP]{
		\includegraphics[width=3cm]{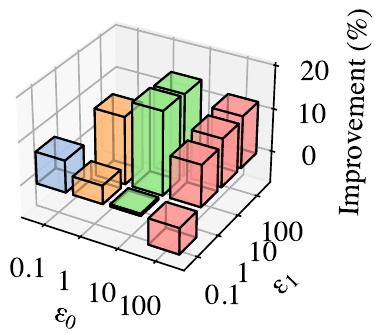}
	}
        \subfigure[TAG-BBBP]{
		\includegraphics[width=3cm]{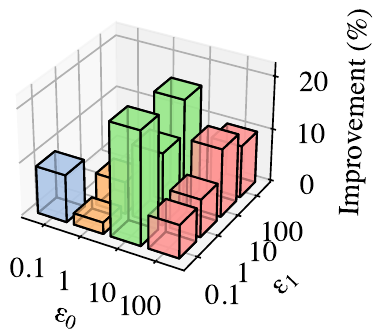}
	}
 
	\subfigure[GCN-Tox21]{
		\includegraphics[width=3cm]{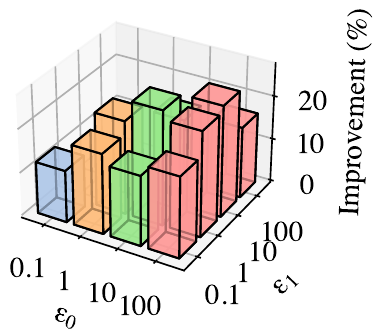}
	}
	\subfigure[kGNNs-Tox21]{
		\includegraphics[width=3cm]{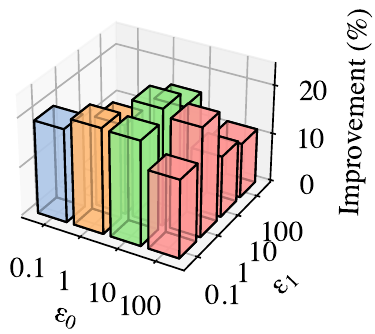}
	}
	\subfigure[TAG-Tox21]{
		\includegraphics[width=3cm]{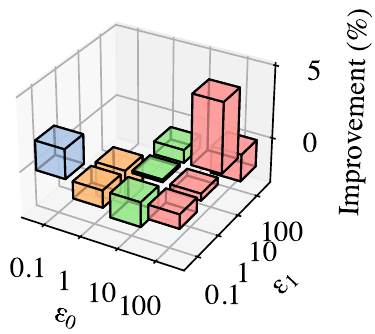}
	}
	\subfigure[LGCN-Tox21]{
		\includegraphics[width=3cm]{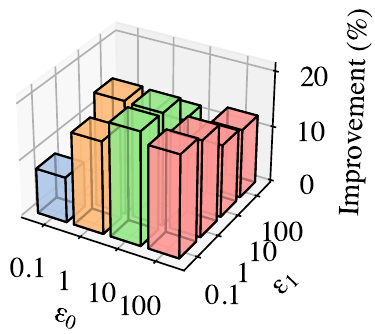}
	}
\caption{We select different privacy budgets for the generation of contrasting pairs to investigate the extent to which the performance decrease caused by the DP mechanism can be alleviated. The figures above illustrate the improvements achieved with different combinations of privacy budgets.}
	\label{fig:pb}
\end{figure*}

The third experiment concerns the privacy budget $\epsilon$, which controls how much privacy leakage can be tolerated. In other words, the privacy budget $\epsilon$ determines how much noise will be introduced. Note that to obtain a pair of positive contrasting views, we need to augment the original graph twice, with privacy budgets $\epsilon_0$ and $\epsilon_1$. In this experiment, we focus on studying the performance of FGCL with $\epsilon_0\neq\epsilon_1$ instead of $\epsilon_0=\epsilon_1$. The values of the privacy budget are selected from $[0.1, 1, 10, 100]$. Therefore, there are $10$ possible combinations on each dataset for a selected base model. Figure \ref{fig:pb} shows the experimental results. We notice that the overall performances are similar to those in Table \ref{tab:comparison}. However, careful fine-tuning of this hyperparameter can yield better outcomes than the results listed in Table \ref{tab:comparison}, even on the ClinTox dataset. Nevertheless, there is no regular pattern indicating how to choose an optimal combination of $\epsilon_0$ and $\epsilon_1$. Hence, this hyperparameter should be selected in accordance with the privacy protection requirements.

\section{Related Work}
In this section, we summarize the research progress on two topics related to this paper: graph contrastive learning and federated graph learning.

\subsection{Graph Contrastive Learning}
Graph contrastive learning (GCL) has emerged as a delicate tool for graph representation learning. DGI \cite{dgi}, one of the first research works to introduce the concept of contrastive learning \cite{cv-contra-1} into the graph learning domain, adopts mutual information maximization as the learning objective to conduct contrastive learning between a graph and a corrupted instance thereof. Subsequently, other researchers have borrowed the same idea with different contrasting samples to conduct contrastive learning. For example, GCC \cite{gcc} selects different graph instances from different datasets to construct contrasting pairs, GraphCL uses graph augmentation methods for this purpose, and MVGRL \cite{mvgrl} and DSGC \cite{dsgc} generate multiple views to serve as the contrasting pairs. The success of GCL can be seen from its broad scope of application in real-world scenarios, including recommender systems \cite{rl-rec} like \cite{hmg-cr, simplegcl} and smart medicine or health service \cite{geomgcl, fl-privacy-health}.

\subsection{Federated Graph Learning}
Federated graph learning (FGL) is a cross-disciplinary field lying at the intersection of graph neural networks (GNNs) and federated learning (FL). It leverages the advantages of FL to address various limitations existing in the graph learning domain and has achieve great success in many scenarios. A representative application case of FGL is molecular learning \cite{fedgraphnn}, in which FGL can help diverse institutions efficiently collaborate to train models based on the small-molecule graphs stored at each institution without transferring their classified data to a centralized server \cite{molecular-1, molecular-2}. Moreover, FGL is also applied in recommender systems \cite{dp-privrec, fedgnn}, social network analysis \cite{fed-social}, and the Internet of Things \cite{asfgnn}. Various toolkits are available to help researchers quickly build their own FGL models, such as TensorFlow Federated\footnote{https://www.tensorFlow.org/federated} and PySyft \cite{pysyft}. However, these toolkits do not provide graph datasets, benchmarks, or high-level APIs for implementing FGL. He \etal \cite{fedml, fedgraphnn} developed an FGL-focused framework named \textit{FedGraphNN}, which is used in this paper. This framework provides comprehensive and high-quality graph datasets, convenient and high-level APIs, and tailored graph learning settings to facilitate research regarding FGL.

\section{Conclusion}
This paper proposes a novel federated graph contrastive learning method named FGCL, which is the first work on privacy-aware graph contrastive learning in federated scenarios. Inspired by our observation of the similarity between differential privacy on graph edges and graph augmentation in graph contrastive learning, we innovatively adopt graph contrastive learning methods to help a model achieve robustness against the noise introduced by the DP mechanism. According to comprehensive experimental results, the proposed FGCL method alleviates the performance decrease caused by the noise introduced by the DP mechanism.

\section*{Acknowledgments}
This research work was supported by the Australian Research Council (ARC) under Grant Nos. DP220103717, LE220100078, LP170100891 and DP200101374 and was partially supported by the APRC - CityU New Research Initiatives (No. 9610565, Start-up Grant for the New Faculty of the City University of Hong Kong), two SIRG - CityU Strategic Interdisciplinary Research Grants (No. 7020046 and No. 7020074), and the CCF-Tencent Open Fund.

\bibliographystyle{elsarticle-harv}
\bibliography{ref}

\end{document}